\pdfoutput=1
\documentclass[11pt]{article}

\PassOptionsToPackage{table}{xcolor}

\usepackage{acl}
\usepackage{times}
\usepackage{latexsym}
\usepackage[T1]{fontenc}
\usepackage[utf8]{inputenc}
\usepackage{microtype}
\usepackage{inconsolata}

\usepackage{graphicx}
\usepackage{booktabs}
\usepackage{graphicx}
\definecolor{before}{HTML}{F9EEED}
\definecolor{after}{HTML}{EDEAFB}

\title{\textsc{Palo}: A Polyglot Large Multimodal Model for 5B People}

\author{%
  Muhammad Maaz$^{1}$\textsuperscript{\textnormal{*}}, Hanoona Rasheed$^{1}$\textsuperscript{\textnormal{*}}, Abdelrahman Shaker$^{1}$, Salman Khan$^{1,2}$ \\
  \textbf{Hisham Cholakal}$^{1}$, \textbf{Rao M. Anwer}$^{1,3}$, \textbf{Tim Baldwin}$^{1,4}$, \textbf{Michael Felsberg}$^{5}$, \textbf{Fahad S. Khan}$^{1,5}$\\
  [0.25cm]
  {\fontsize{10.5pt}{12pt}\selectfont $^{1}$Mohamed bin Zayed University of AI, $^{2}$Australian National University, $^{3}$Aalto University}\\
 {\fontsize{10.5pt}{12pt}\selectfont $^{4}$The University of Melbourne, $^{5}$Linköping University}\\
  \hypersetup{urlcolor=black}
}

\begin{document}
\maketitle
\begin{abstract}
In pursuit of more inclusive Vision-Language Models (VLMs), this study introduces a Large Multilingual Multimodal Model called \textsc{Palo}. \textsc{Palo} offers visual reasoning capabilities in 10 major languages, including English, Chinese, Hindi, Spanish, French, Arabic, Bengali, Russian, Urdu, and Japanese, that span a total of $\sim$5B people (65\% of the world population). Our approach involves a semi-automated translation approach to adapt the multimodal instruction dataset from English to the target languages using a fine-tuned Large Language Model, thereby ensuring high linguistic fidelity while allowing scalability due to minimal manual effort. 
The incorporation of diverse instruction sets helps us boost overall performance across multiple languages especially those that are underrepresented like Hindi, Arabic, Bengali, and Urdu. The resulting models are trained across three scales (1.7B, 7B and 13B parameters) to show the generalization and scalability where we observe substantial improvements compared to strong baselines. We also propose the first multilingual multimodal benchmark for the forthcoming approaches to evaluate their vision-language reasoning capabilities across languages. Code: \url{https://github.com/mbzuai-oryx/PALO}.
\footnotetext[1]{Equally contributing first authors.}
\end{abstract}

\section{Introduction}
\label{sec:intro}

\begin{figure}[!t]
  \centering
    \includegraphics[width=0.98\linewidth]{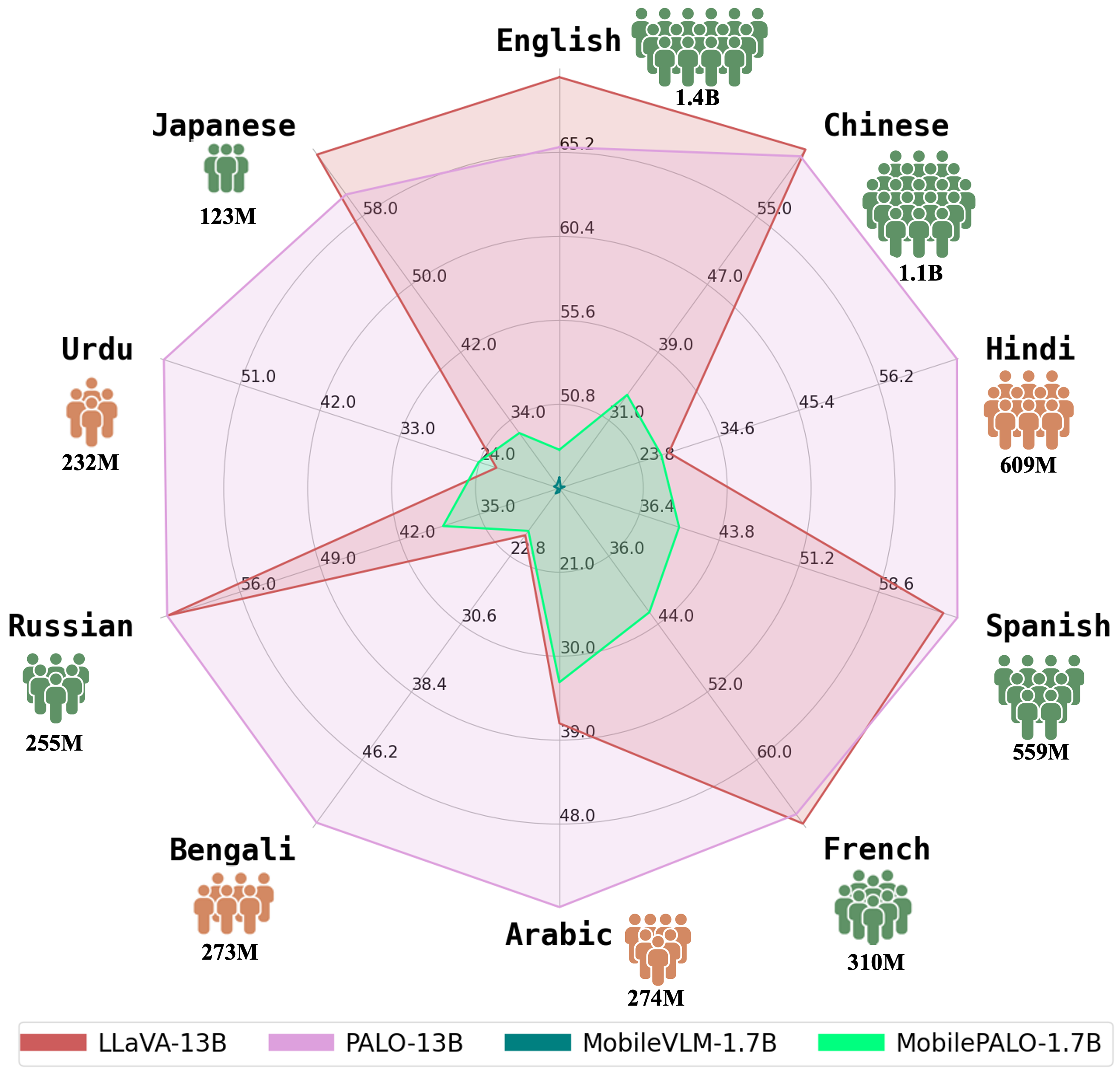}\vspace{-0.8em}
    \caption{\textbf{\textsc{Palo} vs. English-VLMs.} The plot compares \textsc{Palo} with corresponding Vision-Language Models (VLMs) across 10 different languages. These languages include English, Chinese, Hindi, Spanish, French, Arabic, Bengali, Russian, Urdu, and Japanese, collectively covering approximately 5B people and 65\% of the global population. English-trained VLMs, such as LLaVA and MobileVLM, exhibit poor performance on low-resource languages including Hindi, Arabic, Bengali, and Urdu, due to the under-representation of these languages during their training phases. \textsc{Palo}, in contrast, is a unified model that can hold conversations simultaneously in all the ten languages, demonstrating consistent performance across the board.}
    \label{fig:fig}
\vspace{-1em}
\end{figure}

Propelled by advancements in generative AI, Large Multimodal Models (LMMs) \cite{liu2023llava, zhu2023minigpt, dai2023instructblip} have emerged as a pivotal advancement in the field, seamlessly bridging the gap between vision and language tasks. While initial efforts such as LLaVA~\cite{liu2023llava} and miniGPT4 \cite{zhu2023minigpt} have demonstrated intriguing performance in synthesizing effective textual responses based on visual inputs, they have predominantly focused on English, leaving a significant gap in multimodal understanding for non-English languages. As a result, the existing LMMs generally overlook the linguistic diversity of the global population, particularly languages spoken by large groups, such as Chinese, Hindi, Spanish, French, Arabic, Bengali, Russian, Urdu, and Japanese, which collectively account for billions of native speakers. Our work addresses this disparity by developing the first fully open-source multilingual LMM called \textsc{Palo}, which encompasses ten major languages covering 65\% of the global population, with a special focus on languages underrepresented in the current multimodal models.

The challenge lies in the scarcity of high-quality multilingual multimodal data compared to English. Addressing the challenge of limited high-quality data, especially for under-represented languages such as Hindi, Arabic, Bengali, and Urdu, our approach involves careful analysis and subsequent refinement of translations produced by a state-of-the-art Large Language Model (LLM)~\cite{brown2020language} for each target language. By identifying and correcting translation inaccuracies through human intervention, we generate a high-quality multilingual dataset. This curated dataset then serves as the foundation for refining the target language annotations, ensuring a more accurate and nuanced representation of the target language in training.

Leveraging our high-quality multilingual vision-language instruction dataset and the recent advances in large multimodal modeling, we develop \textsc{Palo} as a \emph{unified} model that can simultaneously answer questions in ten different languages. Our training pipeline offers substantial gains in low-resource languages (underrepresented in the LLM training datasets) while maintaining (or further improving) performance on high-resource languages. The contributions of this work are as follows,
\begin{itemize}\setlength{\itemsep}{0mm}
\item We develop \textsc{Palo}: the first multilingual Large Multimodal Model (LMM) covering ten major languages, facilitating vision-language reasoning through a generic model capable of generating responses in any of the ten languages.
\item We assemble an extensive multilingual (10 languages) instruction-tuning dataset, through a critical analysis and subsequent refinement of a state-of-the-art Large Language Model’s target language translations. This dataset is pivotal in improving proficiency in processing and generating content that is linguistically precise across multiple languages.
% \item We enhance the performance of an LMM~\cite{liu2023llava} model across diverse language tasks with substantial improvements in understanding and generating content for low-resource languages, without compromising its high-performance capabilities in high-resource languages highlighting the success of our multilingual dataset and fine-tuning approach.
\item We enhance the multilingual performance of state-of-the-art LMMs~\cite{liu2023llava,chu2023mobilevlm}  across three distinct scales i.e., 1.7B, 7B, and 13B parameters to demonstrate the scalability of our training pipeline. The resulting polyglot LMMs demonstrate performance gains on diverse language tasks with substantial improvements in understanding and generating content for low-resource languages, e.g., Hindi, Arabic, Bengali, and Urdu, without compromising its high-performance on high-resource languages e.g., English, Chinese, French, and Spanish.
\end{itemize}

\begin{figure*}[!t]
  \centering
    \includegraphics[width=0.99\linewidth]{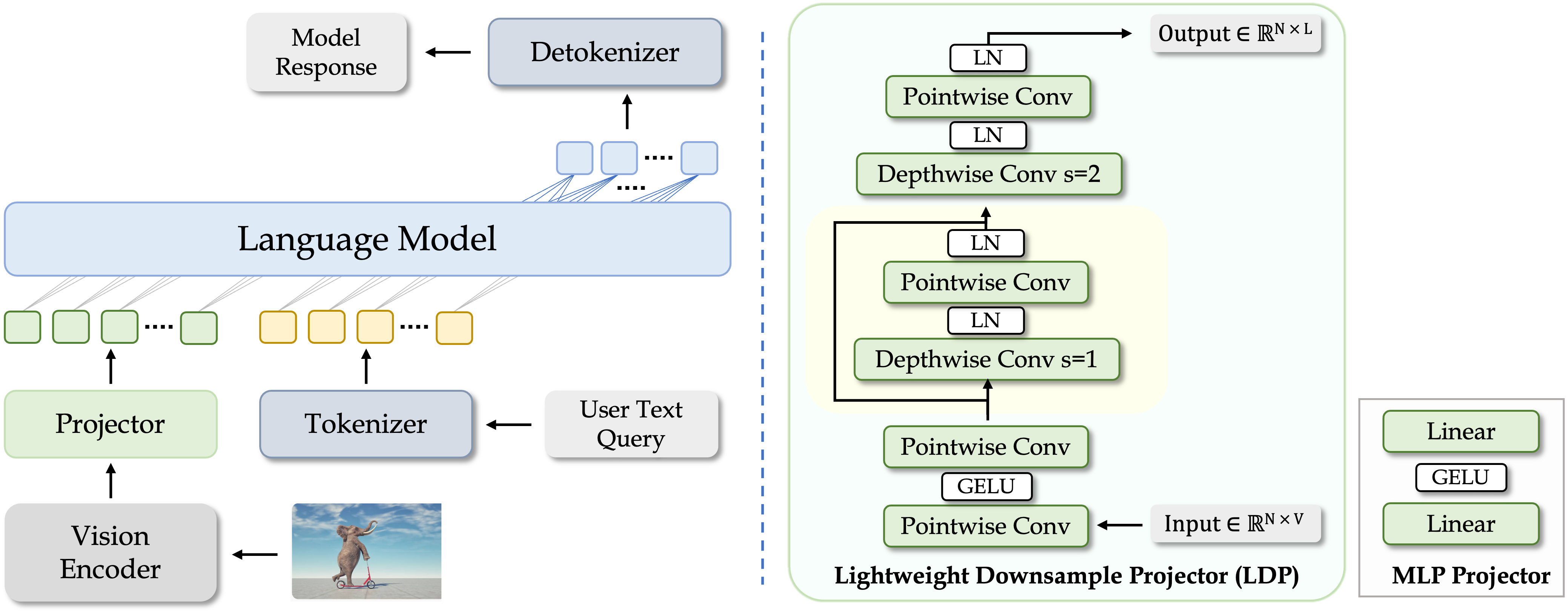}\vspace{-0.8em}
    \caption{\textbf{Architecture overview of \textsc{Palo}.} (\emph{left}) The model consists of a vision encoder that encodes the image, followed by a projector that projects the vision features into the input embedding space of the language model. The user's text query is tokenized, and the tokens are concatenated with the vision tokens before being input into the causal language model to generate the response. 
    For the \textsc{Palo} 7B and 13B variants, Vicuna is used as the Large Language Model while MobileLLaMA  \cite{chu2023mobilevlm} is used as the Small Language Model in our Mobile\textsc{Palo}-1.7B variant. CLIP ViT-L/336px is used as the vision encoder in all variants.
    (\emph{right}) Projectors used in different variants of \textsc{Palo} are shown. For the \textsc{Palo} 7B and 13B, following~\cite{liu2023llava}, we use a two-layer MLP projector with GELU activation. For our mobile version of \textsc{Palo} (Mobile\textsc{Palo}-1.7B), we use a Lightweight Downsample Projector (LDP) from~\cite{chu2023mobilevlm}. It utilizes depth-wise separable convolutions to downsample the image tokens, making it faster than a standard MLP projector.}
    \label{fig:blockdiagram}
\vspace{-0.5em}
\end{figure*}

\section{Related Works}
The introduction of Large Language Models (LLMs) has significantly advanced the field of natural language processing. However, the development of multilingual LLMs has faced considerable challenges, primarily due to the skewed distribution of language data~\cite{costa2022no}. English and European languages dominate existing datasets, leaving widely spoken languages such as Mandarin Chinese and Hindi underrepresented~\cite{eberhard2015ethnologue}. Moreover, integrating multiple languages into LLMs often leads to a decline in English language performance~\cite{scao2022language}, highlighting a major challenge in maintaining cross-lingual performance.

Recent efforts have aimed to address these challenges by developing multilingual LLMs with enhanced capabilities~\cite{almazrouei2023falcon, llama-2, le2211bloom, wei2023polylm}. BLOOM~\cite{le2211bloom}, trained on the ROOTS corpus~\cite{laurenccon2022bigscience} that comprises sources in 46 languages, marks a substantial step forward in making LLMs accessible across a wide range of languages, including those with fewer resources.
% PolyLM~\cite{wei2023polylm} introduces bilingual data integration, curriculum learning, and a multilingual self-instruct method, outperforming other open-source models in multilingual tasks without compromising English performance. 
PaLM~\cite{chowdhery2023palm} showcases the advantages of scaling, achieving improved results in both monolingual and multilingual tasks through sophisticated training techniques and a novel pathways architecture.

Advancements in Large Multimodal Models (LMMs) have evolved from basic image-level interactions~\cite{liu2023llava, chu2023mobilevlm} to offering flexibility by focusing on region-specific analysis~\cite{hanoona2023GLaMM} and spatio-temporal conversations~\cite{Maaz2023VideoChatGPT,lin2023video}, highlighting the significant progress in this domain. However, the exploration of multilingual capabilities has been limited. Qwen~\cite{bai2023qwen} and mPLUG-Owl~\cite{ye2023mplug} extend LMM functionalities to process visual inputs in both English and Chinese, showcasing its adaptability in processing bilingual visual information. Ziya-Visual~\cite{lu2023ziya} demonstrates the translation of English image-text datasets into Chinese, employing in-context learning for instruction-response generation. However, these LMMs remain limited to two languages. 

We introduce \textsc{Palo}, the first fully open-source LMM, offering visual reasoning capabilities across ten major languages, addressing the gap in multilingual LMMs. In contrast to GPT-4~\cite{achiam2023gpt} which is closed-source and only accessible via APIs, ours is the largest effort in the open-source domain to extend LMM capabilities to multiple languages.

%\vspace{-0.5em}
\section{\textsc{Palo}: A Polyglot LMM}
\label{sec:method}
Towards more globally accessible Vision-Language Models (VLMs), our model \textsc{Palo} (\textbf{P}olyglot L\textbf{a}rge Mu\textbf{l}timodal M\textbf{o}del) is designed to comprehend and generate content in ten major languages, serving an audience that spans nearly two-thirds of the global population. The architecture of \textsc{Palo} is derived from LLaVA (Large Language and Vision Assistant)~\cite{liu2023llava,liu2023improvedllava} for our larger-scale models (7/13B), and from MobileVLM for our mobile-efficient model (1.7B), ensuring that \textsc{Palo} remains versatile across different computational settings.

The architecture seamlessly integrates a vision encoder with a language model (see Figure~\ref{fig:blockdiagram}). Given an input image and user text query, the model generates an accurate natural language response.

\textsc{Palo} uses CLIP ViT-L/14~\cite{radford2021learning} as the vision encoder followed by a projector to transform vision tokens to the input embedding space of the language model. Following LLaVA~\cite{liu2023llava}, we use a two-layer MLP with GELU activation as the projector for our 7/13B models. However, a lightweight downsample projector (LDP)~\cite{chu2023mobilevlm} is used for Mobile\textsc{Palo}-1.7B model. LDP utilizes depth-wise separable convolutions to downsample the vision tokens, largely reducing the input tokens to the language model and hence significantly reducing the training and inference time. Further, convolutions in LDP have fewer parameters as compared to MLP, making our mobile model both parameter and compute-efficient. The projector used in the different PALO versions are shown in Figure~\ref{fig:blockdiagram}.

The projected vision tokens are then concatenated with the tokenized user text query and passed to the language model for generating the response. As \textsc{Palo} trains on ten languages using an extensive multi-modal instruction tuning dataset, this not only enables more effective utilization of the tokenizer's capacity but also expands the search space, providing a richer context and more challenging examples for training.
the language model. 
This approach significantly enhances the ability of the model to understand and generate responses across a diverse set of languages.

We use Vicuna~\cite{vicuna} as the large language model (LLM) in our 7/13B models and MobileLLaMA~\cite{chu2023mobilevlm} as the small language model (SLM) in Mobile\textsc{Palo}-1.7B model. Vicuna fine-tunes LLaMA-2 on user-shared conversations collected from ShareGPT, while LLaMA-2 is pre-trained on 2T tokens collected from different public sources~\cite{llama-2}. On the other hand, MobileLLaMA performs pretraining on 1.3T tokens from RedPajama-v1~\cite{together2023redpajama} followed by fine-tuning on a publicly available version of ShareGPT data~\cite{huggingface_sharegpt}.

\subsection{Dataset}
\label{sec:dataset}
The primary contribution of our work lies in the meticulous preparation of a comprehensive multilingual vision-language instruction-tuning dataset. We begin by selecting a state-of-the-art LMM model~\cite{liu2023llava} for our focus. To tailor the instruction-tuning dataset more effectively for multiple languages in a scalable way, we leverage an LLM model~\cite{brown2020language} to develop a semi-automated translation pipeline. This approach involves translating the English dataset into the target languages, 
% using our scalable pipeline
thereby creating a robust multilingual dataset, which significantly broadens the linguistic scope and applicability of the model.

%%%%%%%%%%%%%%%%%%%%%%%%%%%%%
% \begin{table}[ht!]
\begin{table*}[t!]
\centering
% \resizebox{0.99\columnwidth}{!}{
\resizebox{\textwidth}{!}{
\begin{tabular}{llc*{8}{c}|cc|c}
\toprule
\textbf{Model} & \textbf{Eng.} & \textbf{Chinese} & \textbf{French} & \textbf{Spanish} & \textbf{Russ.} & \textbf{Japan.} & \textbf{Arabic} & \textbf{Hindi} & \textbf{Bengali} & \textbf{Urdu} & \textbf{Avg.H} & \textbf{Avg.L} & \textbf{Avg.} \\
\midrule

LLaVA-7B & \textbf{67.9} & \textbf{55.7} & \textbf{62.4} & \textbf{64.5} & 55.3 & \textbf{59.2} & 38.9 & 29.4 & 13.9 & 21.8 & \textbf{60.8} & 26.0 & 46.9 \\
\rowcolor{violet!10} \textsc{Palo}-7B & 64.2 & \textbf{55.7} & 58.3 & 61.0 & \textbf{57.4} & 57.5 & \textbf{57.8} & \textbf{57.6} & \textbf{51.7} & \textbf{55.3} & 59.0 & \textbf{55.6} & \textbf{57.7} \\
\midrule
& \color{blue}{-3.7} & \color{blue}{0.0} & \color{blue}{-4.1} & \color{blue}{-3.5} & \color{blue}{+2.1} & \color{blue}{-1.7} & \color{blue}{+18.9} & \color{blue}{+28.2} & \color{blue}{+37.8} & \color{blue}{+33.5} & \color{blue}{-1.8} & \color{blue}{+29.6} & \color{blue}{+10.8} \\
\midrule
LLaVA-13B & \textbf{69.5} & \textbf{62.9} & \textbf{67.5} & 64.6 & 62.3 & \textbf{65.3} & 37.2 & 27.8 & 20.4 & 22.1 & \textbf{65.4} & 26.9 & 49.9 \\
\rowcolor{violet!10} \textsc{Palo}-13B & 65.5 & 62.1 & 66.4 & \textbf{65.9} & \textbf{62.4} & 60.6 & \textbf{56.9} & \textbf{66.8} & \textbf{53.5} & \textbf{59.6} & 63.8 & \textbf{59.2} & \textbf{61.9} \\
\midrule
& \color{blue}{-4.0} & \color{blue}{-0.8} & \color{blue}{-1.1} & \color{blue}{+1.3} & \color{blue}{+0.1} & \color{blue}{-4.7} & \color{blue}{+19.7} & \color{blue}{+39.0} & \color{blue}{+33.1} & \color{blue}{+37.5} & \color{blue}{-1.5} & \color{blue}{+32.3} & \color{blue}{+12.0} \\
\midrule
MobileVLM-1.7B & 46.6 & 23.2 & 28.1 & 29.1 & 28.1 & 26.4 & 12.4 & 13.7 & 15.6 & 15.6 & 30.3 & 14.3 & 23.9 \\
\rowcolor{violet!10} Mobile\textsc{Palo}-1.7B & \textbf{48.2} & \textbf{34.0} & \textbf{42.6} & \textbf{40.1} & \textbf{38.2} & \textbf{32.5} & \textbf{32.8} & \textbf{26.8} & \textbf{19.9} & \textbf{24.1} & \textbf{39.3} & \textbf{25.9} & \textbf{33.9} \\
\midrule
& \color{blue}{+1.6} & \color{blue}{+10.8} & \color{blue}{+14.5} & \color{blue}{+11.0} & \color{blue}{+10.1} & \color{blue}{+6.1} & \color{blue}{+20.4} & \color{blue}{+13.1} & \color{blue}{+4.3} & \color{blue}{+8.5} & \color{blue}{+9.0} & \color{blue}{+11.6} & \color{blue}{+10.0} \\
\bottomrule
\end{tabular}
}\vspace{-0.5em}
\caption{\textbf{Standard VLMs vs \textsc{Palo} on multi-lingual multimodal evaluation.} The table shows the comparison of LLaVA and MobileVLM with \textsc{Palo} on ten languages on the specially adapted multilingual version of LLaVA-Bench (In-the-Wild). LLaVA 7/13B and MobileVLM-1.7B are fine-tuned on LLaVA-Instruct-665K, and \textsc{Palo} is fine-tuned on LLaVA-Instruct-665K plus the LLaVA-Instruct-150K translated in all ten languages. All models are pretrained on CC-595K~\cite{liu2023llava} dataset. Avg.H and Avg.L represent the average over high-resource (English, Chinese, French, Spanish, Russian and Japanese) and low-resource (Arabic, Hindi, Bengali and Urdu) languages respectively. Avg. represents the average over all the languages.}
\label{results_table2}
\end{table*}
% \end{table}
%%%%%%%%%%%%%%%%%%%%%%%%%%%%%

\begin{figure}[!t]
  \centering
    \includegraphics[width=1.0\linewidth]{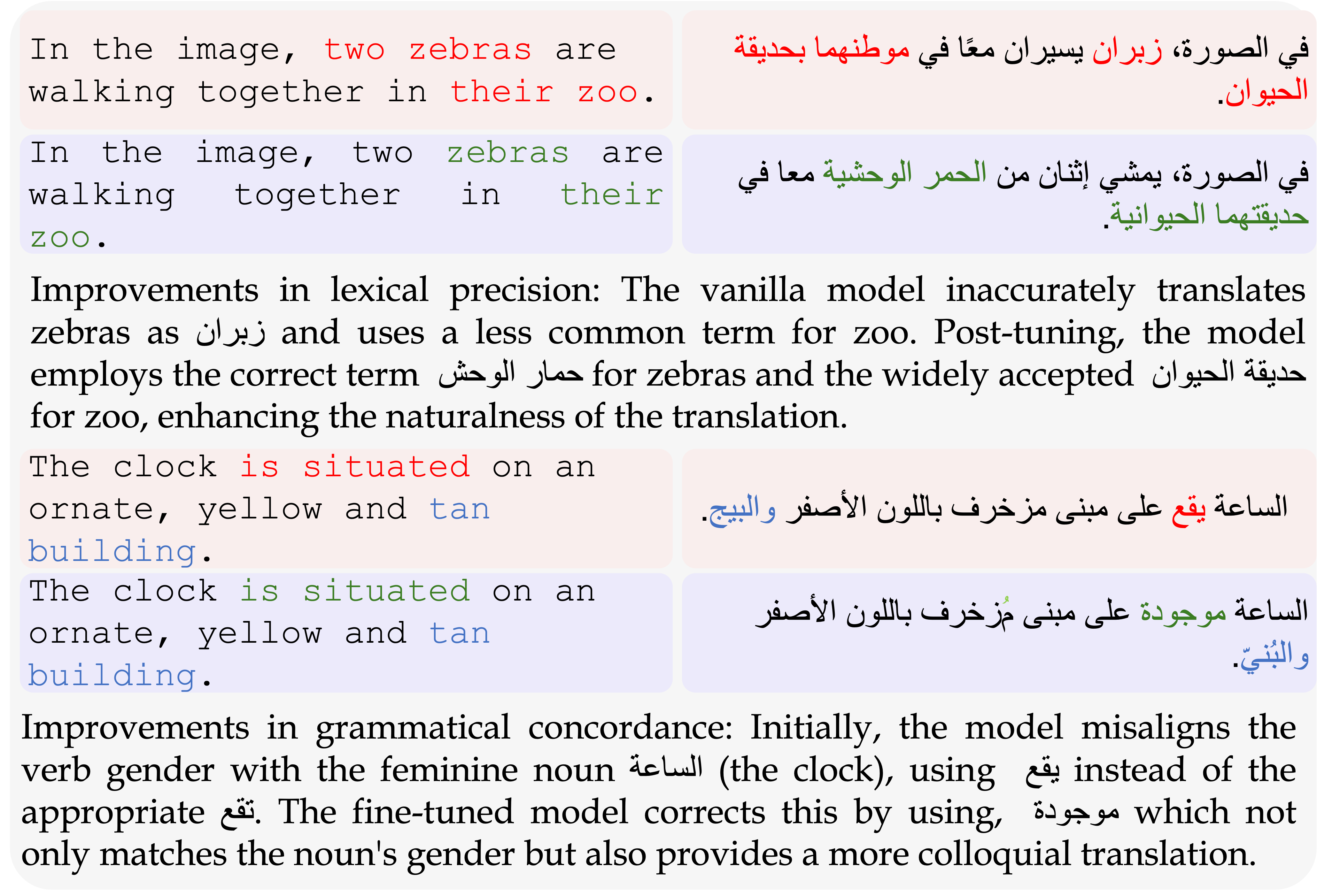}\vspace{-0.8em}
    \caption{\textbf{Qualitative results showing the impact of fine-tuning}. 
    Comparative visualization of English to Arabic translations \colorbox{before}{before} and \colorbox{after}{after} fine-tuning the LLM.
    The figure shows improvements in language-specific issues such as accurate vocabulary usage, gender agreement, and grammatical correctness, 
    highlighting the enhanced performance of the fine-tuned model.
    }
    \label{fig:data_vis}
\vspace{-1em}
\end{figure}

\noindent \textbf{Translation Process and Challenges}:
A naive translation approach from English to the target languages using an LLM model~\cite{brown2020language} effectively conveys the basic meanings but introduces several linguistic challenges specific to each language. Issues such as punctuation, grammatical nuances, translation consistencies, and gender usage errors are observed via a direct LLM-based translation (refer Figure.\ref{fig:data_vis}). These challenges vary greatly due to the linguistic diversity of the languages involved, from the tonal complexities of Chinese to the script variances in Hindi and the gender-specific intricacies of languages like Spanish, Arabic and Russian. For instance, in the case of Arabic, common punctuation mistakes involve incorrect spacing around commas and periods. Nunnation, vital in Arabic grammar, is sometimes omitted or wrongly applied. Additionally, certain English words remain untranslated in the translated text, and there are instances where verbs are incorrectly converted to nouns alongside incorrect gender alignment in translations that pose significant concerns, given the gender-specific nature of grammar in some target languages.

\noindent \textbf{Addressing the Challenges}:
To improve the quality of the translated dataset, we employ a combination of automated and manual verification steps. In this semi-automated pipeline, a team of native speakers for each language provides detailed review and correction of a small subset from initial translations, addressing language-specific issues, gender accuracy, and overall linguistic integrity. Automated scripts are tailored for each language to correct common punctuation mistakes and optimize the verification process.

\noindent \textbf{Fine-tuning of the LLM}:
Acknowledging the limitations of the LLM for multilingual translations, we leverage manually verified and corrected translations (1K conversations per language) as a high-quality dataset for fine-tuning the LLM. This fine-tuning is focused not only on improving translation accuracy but also on aligning the outputs with the specific attributes of each language, such as tone and orthography. The enhanced and fine-tuned LLM is then employed to translate the extensive VLM instruction tuning dataset~\cite{liu2023llava} comprising approximately 150K instructions (i.e. LLaVA-Instruct-150K from ~\cite{liu2023llava}) from English into the respective languages. We use GPT3.5-Turbo as the translation model and finetune it using OpenAI finetuning platform. 

\noindent \textbf{Impact of the Refined Dataset}:
This process results in a comprehensive and high-quality multilingual dataset, crucial for the effective fine-tuning of \textsc{Palo}. The improved dataset not only addresses specific aspects of each language but also markedly improves the ability of the model to process and generate contextually relevant and grammatically accurate content in all included languages. For instance, Figure~\ref{fig:data_vis} highlights two key improvements in English to Arabic translation, the first example shows enhanced lexical precision, and the second shows improved grammatical concordance. Integrating this dataset into the LMM's training process is the key to expanding its capabilities to include both English and nine other languages effectively.

\section{Experiments}
\label{sec:experiments}
\subsection{Implementation Details}
Similar to the LLaVA and MobileVLM baselines, we pretrain our models on a subset of CC3M dataset called CC-595K~\cite{liu2023llava}. During pretraining, only the projector is learned and the rest of the model components are kept frozen. We train the model for 1 epoch with an overall batch size of 256 with 32 batch size per GPU on eight A-100 40GB GPUs. The model is optimized using Adam optimizer and cosine LR scheduler with a learning rate of 2e-3. The pertaining takes around 1.5 hours for 1.7B, 5 hours for 7B and almost 9 hours for the 13B model.

We fine-tune our model on a diverse instruction dataset comprising conversations from ten languages. Specifically, 665K instructions from LLaVA-Instruct-665K~\cite{liu2023improvedllava} are used for English, and approximately 150K conversations from LLaVA-Instruct-150K~\cite{liu2023llava} for Chinese, French, Spanish, Russian, Japanese, Arabic, Hindi, Bengali and Urdu, summing up to almost 2.1M instructions in total. During fine-tuning, only the vision encoder is kept froze and the rest of the model is trained. Projector is fully trained while language model is LORA~\cite{hu2022lora} fine-tuned with $\alpha=128$. We train the model for 1 epoch with an overall batch size of 128 with 16 batch size per GPU on eight A-100 GPUs. We use 40GB A-100 GPUs for 1.7/7B variants and 80GB A-100 GPUs for 13B variants. The model is optimized using Adam optimizer and cosine LR scheduler with 2e-5 base learning rate for the projector and 2e-4 for the language model. The finetuning takes around 12 hours for 1.7B, 42 hours for 7B and almost 76 hours for the 13B model.

\subsection{High-resource vs Low-resource Languages}
Our work trains and evaluates on ten languages divided into two groups, high-resource and low-resource languages. English, Chinese, French, Spanish, Russian and Japanese are considered high-resource languages as the language model training data contains a reasonable number of samples from these languages. On the other hand, Arabic, Hindi, Bengali and Urdu are categorized as low-resource languages as they are under-represented in the language model training data.

%%%%%%%%%%%%%%%%%%%%%%%%%%%%%%%%%%%%%%%%%%
% \begin{table}[ht!]
\begin{table*}[t!]
\centering
% \resizebox{0.99\columnwidth}{!}{
\resizebox{0.98\textwidth}{!}{
\begin{tabular}{lcc*{8}{c}|c}
\toprule
\textbf{Data} & \textbf{English} & \textbf{Chinese} & \textbf{French} & \textbf{Spanish} & \textbf{Russian} & \textbf{Japanese} & \textbf{Arabic} & \textbf{Hindi} & \textbf{Bengali} & \textbf{Urdu} & \textbf{Avg.} \\
\midrule

665K-English & \cellcolor{gray!20}\textbf{67.9} & \textbf{55.7} & \textbf{62.4} & \textbf{64.5} & 55.3 & \textbf{59.2} & 38.9 & 29.4 & 13.9 & 21.8 & 46.9 \\
150K-Chinese & 59.3 & \cellcolor{gray!20}55.0 & 60.0 & 57.0 & 32.9 & 40.5 & 21.2 & 20.3 & 21.7 & 19.3 & 38.7 \\
150K-French & 51.0 & 41.0 & \cellcolor{gray!20}57.8 & 54.4 & 35.4 & 54.6 & 17.6 & 23.2 & 13.1 & 16.7 & 36.5 \\
150K-Spanish & 61.1 & 52.2 & 54.8 & \cellcolor{gray!20}61.6 & 50.1 & 51.7 & 27.8 & 24.4 & 15.4 & 18.5 & 41.8 \\
150K-Russian & 55.2 & 51.1 & 62.2 & 60.6 & \cellcolor{gray!20}\textbf{57.8} & 50.9 & 25.3 & 28.2 & 13.6 & 16.7 & 42.2 \\
150K-Japanese & 54.5 & 41.1 & 59.2 & 57.6 & 36.1 & \cellcolor{gray!20}57.6 & 18.0 & 23.6 & 13.3 & 18.4 & 37.9 \\
150K-Arabic & 67.8 & 42.9 & 56.4 & 54.7 & 38.4 & 44.7 & \cellcolor{gray!20}56.0 & 25.7 & 19.4 & 33.4 & 43.9 \\
150K-Hindi & 52.2 & 39.1 & 56.8 & 54.0 & 35.0 & 33.4 & 18.4 & \cellcolor{gray!20}54.1 & 12.8 & 23.8 & 37.9 \\
150K-Bengali & 26.4 & 40.2 & 56.0 & 54.5 & 37.3 & 26.0 & 12.8 & 16.3 & \cellcolor{gray!20}34.8 & 14.0 & 31.8 \\
150K-Urdu & 28.9 & 30.6 & 44.6 & 50.1 & 22.5 & 16.0 & 22.1 & 25.5 & 20.9 & \cellcolor{gray!20}47.7 & 30.9 \\
\rowcolor{violet!10} Combined & 64.2 & \textbf{55.7} & 58.3 & 61.0 & 57.4 & 57.5 & \textbf{57.8} & \textbf{57.6} & \textbf{51.7} & \textbf{55.3} & \textbf{57.7} \\
% \midrule
% 150K-Bengali & 26.4 & 40.2 & 56.0 & 54.5 & 37.3 & 26.0 & 12.8 & 16.3 & 34.8 & 14.0 & 31.8 \\
% 220K-Bengali & \textbf{18.2} & \textbf{18.6} & \textbf{xx.x} & \textbf{46.1} & \textbf{xx.x} & \textbf{xx.x} & \textbf{xx.x} & \textbf{xx.x} & \textbf{38.3} & \textbf{xx.x} & \textbf{xx.x} \\
\bottomrule
\end{tabular}
}\vspace{-0.5em}
\caption{\textbf{Ablation on multi-lingual fine-tuning dataset.} The table shows an effect of performance on ten languages when using fine-tuning data from different languages. Models with 7B parameters are used for this ablation.}
\label{results_table3}
\end{table*}
% \end{table}
%%%%%%%%%%%%%%%%%%%%%%%%%%%%%%%%%%%%%%%%%%

For example, LLaMA-2~\cite{llama-2} pretraining data contains almost 2 trillion tokens, out of which 89.7\% are of English and almost 1.92\% is for Chinese, French, Spanish, Russian, Japanese, and 21 more similar languages. While the representation of Arabic, Hindi, Bengali and Urdu is negligible.
Similarly, MobileLLaMA~\cite{chu2023mobilevlm} pretrains on RedPajama-v1~\cite{together2023redpajama} dataset which consist of almost 1.3 trillion tokens, predominantly English tokens.

\subsection{Results}
In evaluating the multilingual capabilities of VLMs, we conduct a comprehensive evaluation across various languages, utilizing a high-quality evaluation set. This set is constructed by translating the LLaVA-Bench (In-the-Wild)~\cite{liu2023llava} into all target languages using GPT-4-Turbo~\cite{achiam2023gpt}, with particular attention to preserving linguistic authenticity and mitigating common issues of automated translations through careful human correction. The benchmark comprises 24 diverse and challenging images from different domains, such as indoor and outdoor scenes, memes, and artwork, each with detailed descriptions and a set of 60 questions designed to test the understanding and generalization abilities of the model. 

The results in Table~\ref{results_table2} show that \textsc{Palo} obtains robust performance in high-resource languages, as shown by the 7/13B models scoring an average of 59.0 and 63.8 respectively across these languages. This demonstrates that our multilingual extension has been effectively integrated without compromising the original capabilities of the model.
Further, the model shows good performance improvements in low-resource languages, with average scores rising from 26.0 and 26.9 to 55.6 and 59.2 points, for the 7B and 13B models, respectively. 

% These enhancements are particularly notable for languages like Hindi and Urdu, which have seen improvements of +39.0 and +37.5 points for the 13B model, illustrating the profound impact of our tailored dataset and fine-tuning process.

The overall performance across all ten languages also improves, with the 7B model achieving an average score of 57.65, and the 13B model reaching 61.97. 
% The results show that our multilingual model not only captures the intricate grammatical and semantic nuances of each language but also highlights the viability of multilingual datasets in significantly boosting the performance of language models. 
The data reflects that our approach successfully creates a more inclusive, diverse, and high-performing VLM, capable of handling the complex landscape of global languages in vision-language tasks (see Figures~\ref{fig:fig} and \ref{fig:fig2} for qualitative results).

\begin{figure*}[!t]
  \centering
    \includegraphics[width=0.96\linewidth]{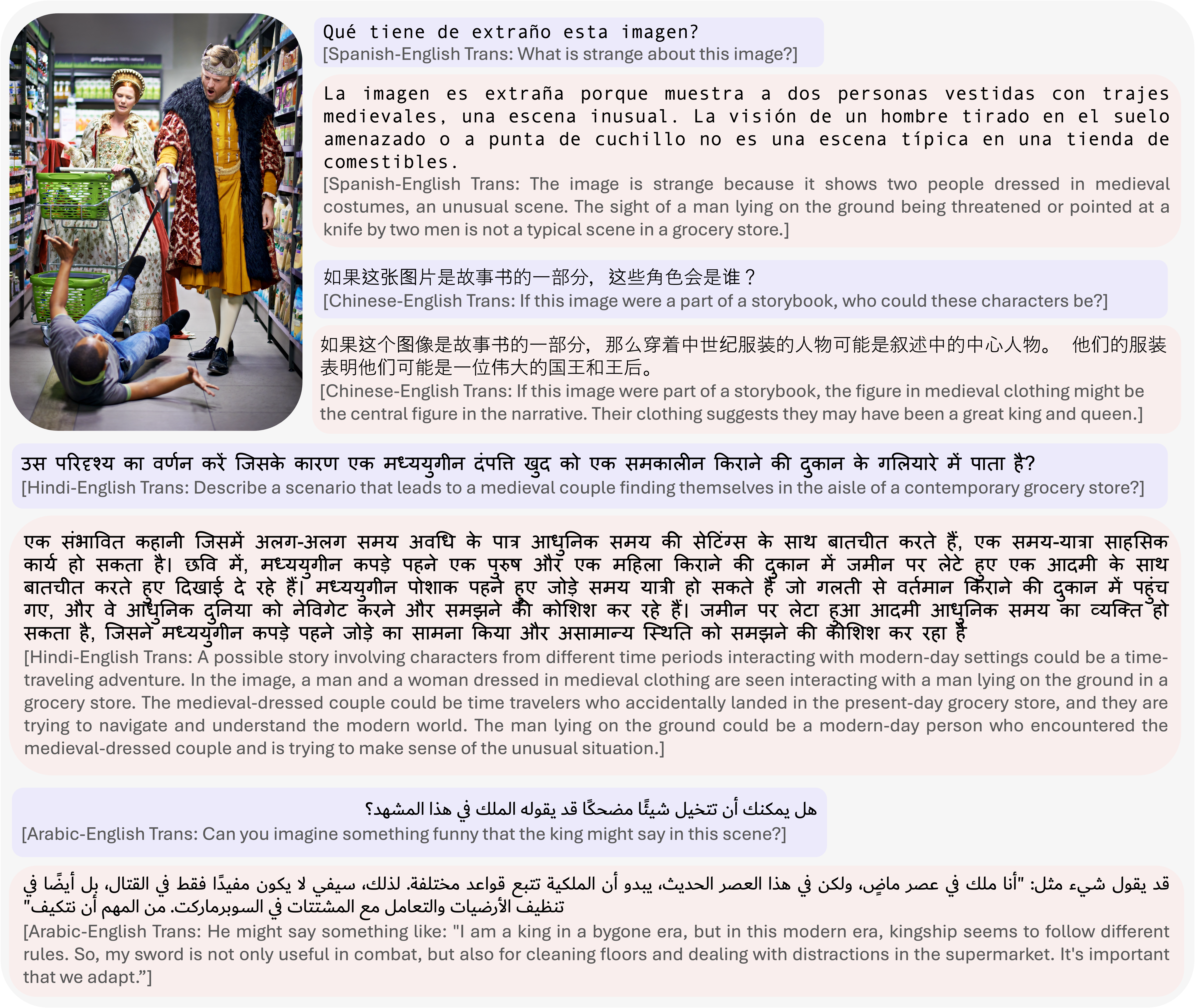}\vspace{-0.8em}
    \caption{\textbf{Qualitative results demonstrating the multilingual capabilities of \textsc{Palo}}. When presented with user queries, the model generates accurate textual responses related to the visual content and the relevant language. The figure highlights its ability to bridge vision and language understanding across diverse languages. In this illustration, we explore dialogues in \textit{two high-resource languages—Spanish and Chinese}—and \textit{two low-resource languages—Hindi and Arabic}. \textsc{Palo} accurately interprets the unusual aspects of an image featuring two individuals in medieval attire within a contemporary supermarket setting. The model exhibits its creative imagination in \textit{Chinese}, proposing a backstory where these characters might be a king and queen from a storybook. In \textit{Hindi}, \textsc{Palo} demonstrates scenario-building by describing a possible situation that brought the medieval couple into the current day as time travellers. At the bottom, \textsc{Palo} displays a touch of humour in \textit{Arabic}, conjuring up a playful dialogue that a king might say, showcasing its subtle understanding of context and culture-specific humour. This image effectively visualizes the advanced ability to process and generate content in multiple languages, reflecting high linguistic precision and cultural intelligence.}
    \label{fig:fig}
\vspace{-1em}
\end{figure*}

\begin{figure*}[!t]
  \centering
    \includegraphics[width=0.98\linewidth]{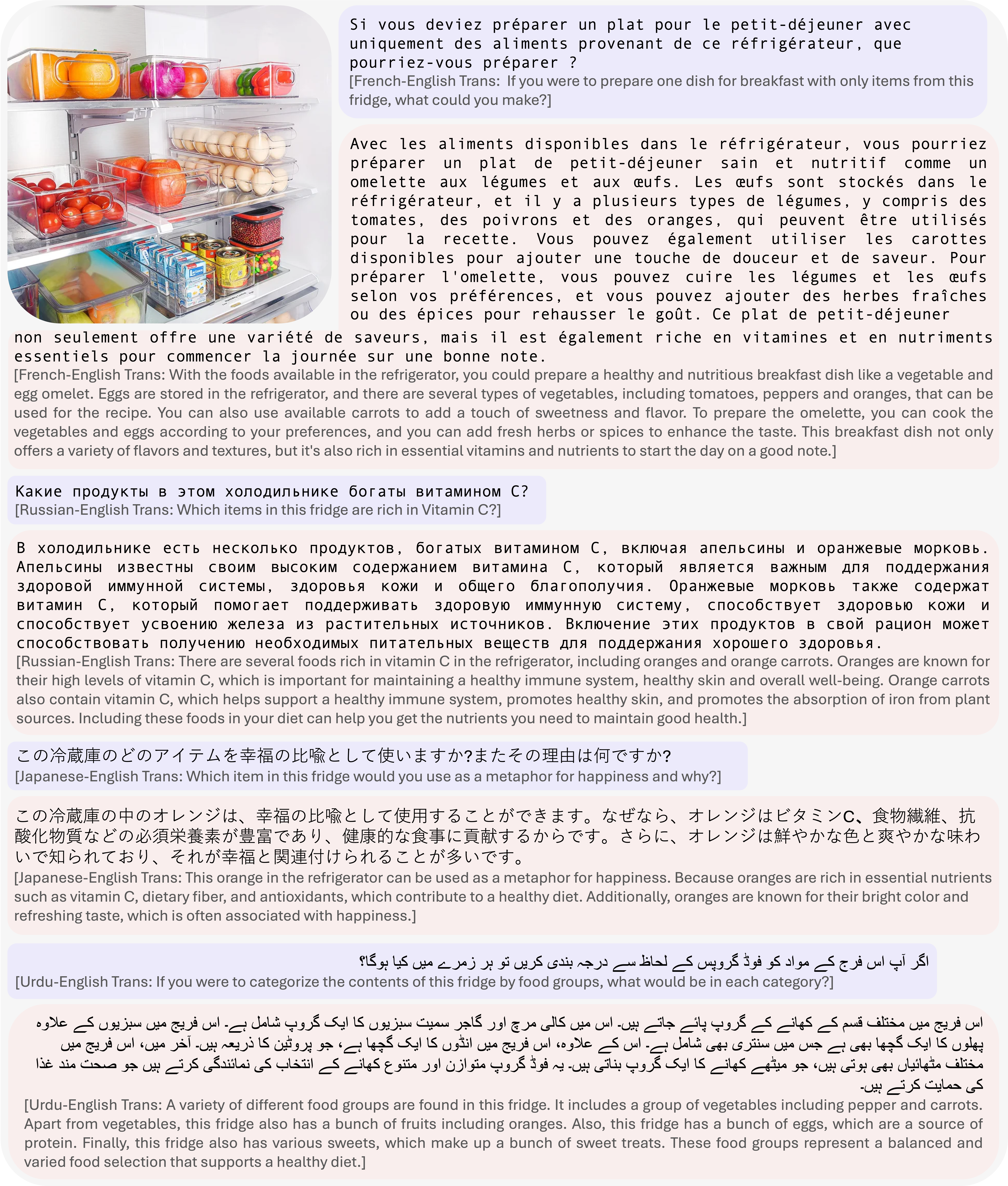}\vspace{-0.8em}
    \caption{\textbf{Qualitative results demonstrating the visual reasoning of \textsc{Palo} and its adeptness in multiple languages}. \textsc{Palo} responds accurately to visual content in a contextually appropriate manner for each language. We illustrate a conversation in \textit{three high-resource languages—French, Russian and Japanese} and \textit{one low-resource language—Urdu}. In the \textit{French} segment, the model shows practical reasoning by suggesting a recipe that utilizes the available ingredients in the fridge, connecting visual perception to culinary suggestions. In \textit{Russian}, \textsc{Palo} identifies items rich in Vitamin C and in the \textit{Urdu} example, the model organizes the fridge contents into food groups, demonstrating its ability to classify items and apply nutritional knowledge. This effectively highlights its ability to switch between languages while maintaining the context of the conversation, reflecting its capacity to generate relevant and culturally aware content in both high-resource and low-resource languages.}
    \label{fig:fig2}
\vspace{-1em}
\end{figure*}

Our mobile model demonstrates consistent improvements across both high-resource and low-resource languages, with an overall average gain of 33.9 points compared to the MobileVLM baseline of 23.9 points. Contrary to the trend observed in the 7/13B model, our mobile version also shows improvements in high-resource languages such as English and Chinese. This performance difference is attributed to the language model pretraining data. LLaMA-2 is trained on 2 trillion tokens with a better representation of high-resource languages compared to MobileLLaMA, which is predominantly trained on 1.3 trillion English tokens. 

\subsection{Ablations}
Table~\ref{results_table3} shows an ablation where we trained our 7B model on 150K translated instructions from each language and evaluated all models across all languages. The results show that the baseline performs better than the language-specific fine-tuned models for high-resource languages, including Chinese, French, Spanish, and Japanese. This is because these languages have less multi-modal data compared to the baseline (i.e., the English model is trained on 665K instructions, while language-specific models are trained on 150K instructions), and due to the noisy semi-automatic translation process. Conversely, the language-specific fine-tuned models perform better in the case of Arabic, Hindi, Bengali, and Urdu, as these languages are under-represented in the LLM pretraining data. Lastly, combined training further improves performance on low-resource languages. 
Further, we found that increasing the quantity of translated multi-modal training data enhances performance. For instance, translating an additional 72K instructions from the GQA dataset~\cite{hudson2019gqa} into Bengali and training with a total of 222K instructions improves Bengali results from 34.8 to 38.3. This study is limited to 150K instructions for each language due to resource constraints.

\section{Conclusion}
\label{sec:conclusion}
We introduce \textsc{Palo}, a polyglot LLM for 5B people, covering almost two-thirds of the world's population. It takes image and user text query as input and effectively converse in both high-resource languages such as English, Chinese, French, Spanish, Russian and Japanese, and low-resource languages such as Arabic, Hindi, Bengali and Urdu. To train our model on ten languages, we translate 150K instructions into each language using custom-tailored LLMs. To fine-tune an LLM on a language-translation task, we use 1K human-annotated conversations for each targeted language. Our final model simultaneously provides competency in ten languages and provides an overall performance improvement on vision-language evaluation. We train \textsc{Palo} across three scales (1.7B, 7B, and 13B) to demonstrate its generalization and scalability across ten languages. Our codes, models, and datasets will be publicly released.

\section{Limitations}
The semi-automated translation process, while efficient, might not fully grasp the deep contextual and cultural nuances inherent to each language. This could impact the capability of the model to comprehend and generate content with the necessary cultural depth, accuracy and precision. Additionally, our selection of ten languages, though it spans two-thirds of the global population, still leaves out a considerable number of the world's languages, indicating room for further expansion to enhance linguistic diversity and inclusivity within VLMs.

\section{Potential Risks}
The use of semi-automated translations could bring forward potential risks tied to biases inherent in LLMs, particularly for low-resource languages. The model must account for nuances in visual data, such as the interpretation of cultural symbols or gestures, to prevent any misrepresentations. The interpretations of the model, influenced by these biases, could lead to inaccuracies in contexts that are culturally sensitive. There is a need to evaluate and adopt necessary training to mitigate such risks.

\section{Use of Data and AI Assistant}
We use LLaVA-Instruct~\cite{liu2023llava} dataset, licensed under Creative Commons Attribution (CCA) 4.0 International, available for use in research. 
Further, the use of GPT models abides by~\cite{openai}. 
Respecting source license information, we will release all datasets created in this work under CCA 4.0 International license.

\section{Human Annotations}
The LLaVA-Bench~\cite{liu2023llava} evaluation for each language is verified and corrected by annotators selected to represent a diverse mix of genders and demographics. Annotators are provided with the English version alongside the translated version. They are given specific instructions to neutralize the tone and biases during the correction process.

\section{Acknowledgements}
The computations were enabled by resources provided by the National Academic Infrastructure for Supercomputing in Sweden (NAISS) at Alvis partially funded by the Swedish Research Council through grant agreement no. 2022-06725, the LUMI supercomputer hosted by CSC (Finland) and the LUMI consortium, and by the Berzelius resource provided by the Knut and Alice Wallenberg Foundation at the National Supercomputer Centre.

% What is given to the annotators
% The detailed process
% How you recruited the annotators
% Demographics of the annotators

% \section{Use of AI Assistants in the Research}
% {\color{red}{The training process of LLaVA involves instruction tuning on machine-generated, multimodal language-image instruction-following data. This enables the model to effectively handle vision-language tasks, as it learns to align the visual and language modalities through instruction tuning. The training process of LLaVA involves instruction tuning on machine-generated, multimodal language-image instruction-following data.}}

\bibliography{custom}

\appendix

% \section{Example Appendix}
% \label{sec:appendix}

% This is an appendix.

\end{document}